%% file: main.tex
\documentclass[hidelinks,10pt,twocolumn,letterpaper]{article}

\usepackage[pagenumbers]{cvpr}              

\def\cvprPaperID{7756} 
\def\confName{CVPR}
\def\confYear{2022}

\usepackage{times}
\usepackage{epsfig}
\usepackage{graphicx}
\usepackage{amsmath}
\usepackage{amssymb}
\usepackage{esint}

\usepackage{booktabs}
\usepackage{multirow}
\usepackage{mathtools}
\usepackage{cite}

\usepackage{pifont}

\newcommand{\cmark}{\ding{51}}%
\newcommand{\xmark}{\ding{55}}%

\usepackage[pagebackref=true]{hyperref}

\newcommand{\R}[0]{\mathbb{R}}

\begin{document}

\title{SurfEmb: Dense and Continuous Correspondence Distributions \\ for Object Pose Estimation with Learnt Surface Embeddings}

\author{%
Rasmus Laurvig Haugaard\\
University of Southern Denmark\\
{\tt\small rlha@mmmi.sdu.dk}
\and
Anders Glent Buch\\
University of Southern Denmark\\
{\tt\small anbu@mmmi.sdu.dk}
}

\maketitle

\begin{abstract}
    We present an approach to learn dense, continuous 2D-3D correspondence distributions over the surface of objects
    from data with no prior knowledge of visual ambiguities like symmetry.
    We also present a new method for 6D pose estimation of rigid objects using the learnt distributions to sample, score and refine pose hypotheses.
    The correspondence distributions are learnt with a contrastive loss, represented in object-specific latent spaces by 
    an encoder-decoder query model and a small fully connected key model.
    Our method is unsupervised with respect to visual ambiguities, yet we show that the query- and key models learn to represent accurate multi-modal surface distributions.
    Our pose estimation method improves the state-of-the-art significantly on the comprehensive BOP Challenge,
    trained purely on synthetic data, even compared with methods trained on real data.
    The project site is at 
    \href{https://surfemb.github.io/}{\color{blue}surfemb.github.io}.
\end{abstract}


\input{sections/intro}
\input{sections/related}
\input{sections/method}
\input{sections/experiments}
\input{sections/limitations}

\input{sections/conclusion}

\section*{Acknowledgements}
The authors would like to thank Thorbjørn Mosekjær Iversen for helpful feedback and discussions.
The project was financially supported by MADE FAST.

{\small
\bibliographystyle{ieee_fullname}
\bibliography{egbib}
}

\end{document}

%% file: sections/intro.tex
\section{Introduction}
Establishing 2D-3D correspondences is a core problem in computer vision.
For an image of an object with no visual ambiguity, a 3D object point can in theory be uniquely identified for a 2D point in the image.
However, under visual ambiguities caused by symmetry, occlusion, lighting, etc., 
there can be a set of possible 3D correspondences.
Ideally, we would have not just a best guess, but a full distribution over possible correspondences.
To the best of our knowledge, we are the first to present such distributions.
We learn the distributions implicitly from data with no information about symmetries or other ambiguities.

\begin{figure}
    \centering
    \includegraphics[width=\linewidth]{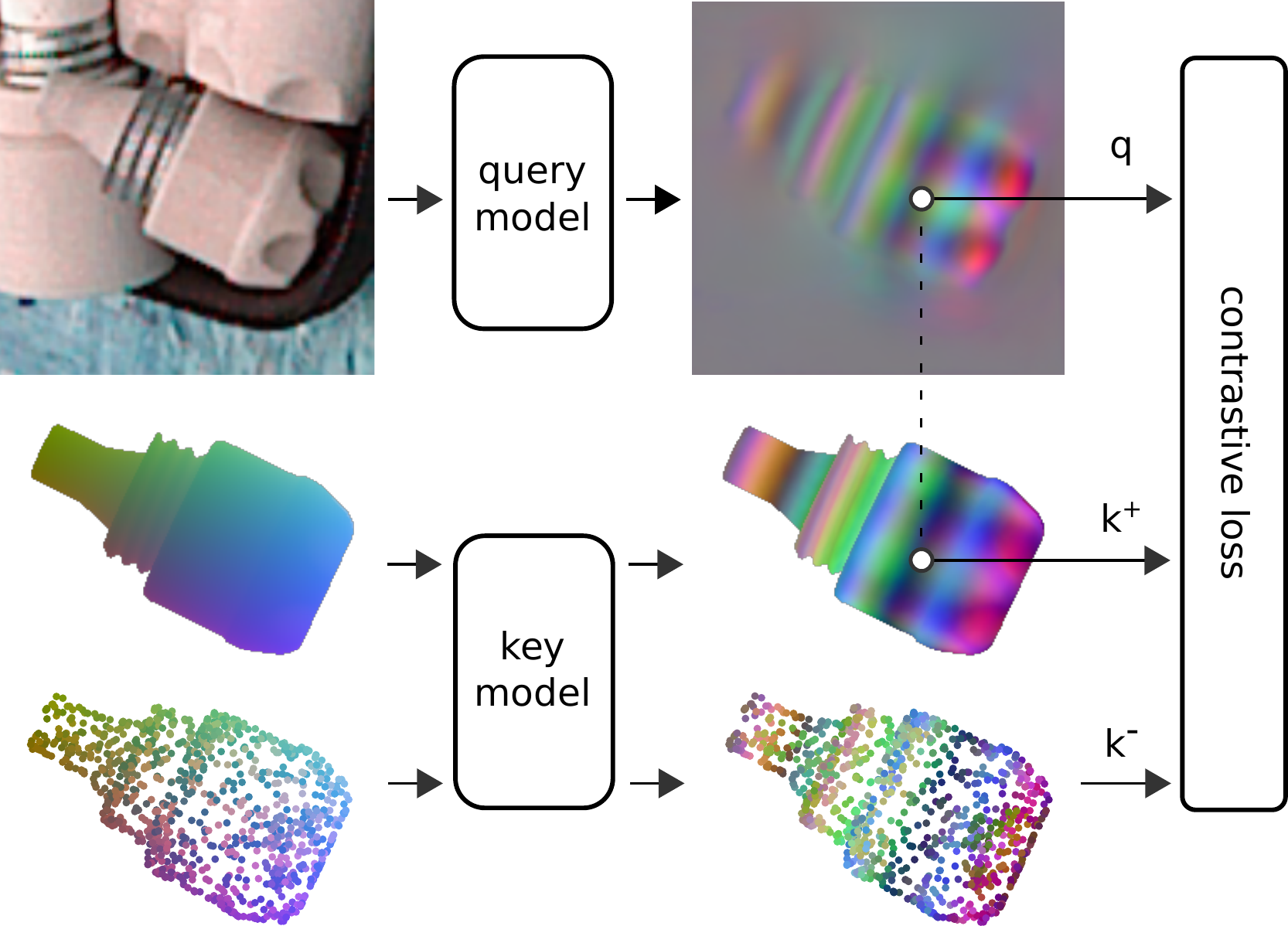}
    \caption{
        Proposed method for learning 2D-3D correspondence distributions.
        An image crop is fed through a query model, providing a query image. 
        Visible object coordinates under the ground truth pose are fed through a key model to provide positive keys, 
        and uniformly sampled object points are fed through the same key model 
        to provide negative keys for contrastive learning.
    }
    \label{fig:training_overview}
\end{figure}

There is a wide range of pose estimation methods for rigid objects.
Many establish 2D-3D correspondences \cite{pix2pose, bb8, pvnet, cdpn, epos, dpodv2} followed by PnP-RANSAC~\cite{ransac}, and mainly differ in \textit{how} they establish correspondences.
Some methods establish correspondences for a fixed set of object keypoints \cite{bb8, pvnet} while others establish dense (pixel-wise) correspondences \cite{pix2pose, cdpn, epos, dpodv2, sopose}.
Regressing coordinates directly \cite{bb8, pix2pose, cdpn, dpodv2, sopose} assumes a uni-modal distribution, 
which is problematic in case of visual ambiguities, 
leading most of them to handle global symmetries explicitly. 
This, however, only handles global symmetries, not other kinds of visual ambiguity.
A recent approach \cite{epos} handles ambiguities implicitly by estimating a probability distribution over surface fragments.
This model is able to represent multi-modal distributions, but the representation is limited by the computational cost of representing a large amount of surface fragments.
They mitigate this by choosing 64 fragments and regressing a within-fragment offset, 
but this still amounts to 256 output channels per object in their encoder-decoder network, 
and their representation is effectively reduced to a discrete probability distribution over 64 refined object coordinates.

Another recent work~\cite{continuous} proposes to establish dense 2D-3D correspondences using continuous surface embeddings with few output channels. However, they only establish a single correspondence per pixel and do not discuss or show the ability to represent distributions or other ways to handle ambiguities, nor do they use it for object pose estimation.

This work presents an approach to learning dense and continuous 2D-3D correspondence distributions with a contrastive loss.
At inference, we evaluate the distributions for approximately 75.000 object coordinates.
The correspondence distributions are represented by two models:
a small fully connected network, the \textit{key model}, mapping object coordinates to key embeddings, and
an encoder-decoder convolutional network, the \textit{query model}, mapping color images to dense query embedding images.
Each query then represents a correspondence distribution over keys, and the two models are trained jointly with a contrastive loss.
This representation enables the models to represent accurate multi-modal correspondence distributions with low computational requirements.

We obtain an initial pose estimate with PnP-RANSAC,
where many pose hypotheses are sampled based on the correspondence distributions and evaluated using a score based on the training loss.
Our models are thus explicitly trained to maximize the score for the correct pose.
We further refine the best scoring pose hypothesis to obtain our final pose estimate.

We evaluate our method on the seven varied datasets in the BOP Challenge~\cite{bop-challenge}.
For ITODD~\cite{itodd}, we show a 79\% relative improvement over the previous state-of-the-art RGB method,
and for T-LESS~\cite{tless} and HB~\cite{hb} our RGB method trained purely on synthetic data is state-of-the-art, even compared with methods trained on real data and methods using depth.
Our RGB method shows state-of-the-art results across all seven datasets, compared to other methods trained on synthetic data,
and our RGBD method trained on purely synthetic data is state-of-the-art on the BOP Challenge, even compared with methods trained on real data.

Our primary contributions are:
\begin{itemize}
    \item Presenting continuous 2D-3D correspondence distributions that are accurate up to ambiguities like symmetry.
    \item A state-of-the-art object pose estimation method using the distributions to sample, score and refine poses, handling symmetries and other ambiguities implicitly.
\end{itemize}

Interactive examples and all code to reproduce the results in this work is available on the project site.

%% file: sections/related.tex
\section{Related Work}



There exists a wide range of pose estimation approaches.
High-performing classical methods \cite{drost, vidal} and hybrids \cite{konig} rely on depth information and use point-pair based voting.

Some learning based approaches \cite{posecnn, deepim, cosypose} regress a representation of the pose directly.
PoseCNN~\cite{posecnn} regresses depth, the projected 2D center and a quaternion for each region-of-interest in a custom detection pipeline.
DeepIM~\cite{deepim} proposes iterative refinement by regressing a pose difference between a render of the pose hypothesis and the input image.
CosyPose~\cite{cosypose} builds on DeepIM, adding several improvements, including 
a continuous rotation parameterization,
handling symmetries explicitly,
and using more recent architectures.
\cite{implicit-orientation-learning, poserbpf} train autoencoders to capture rotation information in a latent representation. 
After training, they build a latent codebook representing a large set of rotations,
and during inference, they either choose the nearest neighbour in the codebook \cite{implicit-orientation-learning} or design a distribution \cite{poserbpf} over rotations.
\cite{multipath} shows that the autoencoder can be shared across multiple objects and generalize to unseen objects.
The latent representation in \cite{implicit-orientation-learning, poserbpf, multipath} handles symmetries implicitly, but assumes that the object frame's origin lies on the planes or axes of symmetries, and does not represent positional ambiguity.
Our method does not assume anything about the placement of the object frame and is able to represent both positional and rotational ambiguities by the distributions of correspondences.

Other learning based approaches are based on establishing 2D-3D correspondences \cite{bb8, pvnet, pix2pose, cdpn, epos, sopose, dpodv2} followed by a variant of PnP-RANSAC.
BB8~\cite{bb8} regresses the 2D coordinates of projected 3D bounding box corners and handles symmetries explicitly by limiting the ground truth pose space based on the object symmetry.
PVNet~\cite{pvnet} regresses vector fields toward the 2D projections of a set of fixed 3D key points and handles symmetries like BB8.
Pix2Pose~\cite{pix2pose} employs an encoder-decoder network to regress dense object coordinates of the object surface.
To handle symmetry, they calculate the loss with respect to the pose that has the smallest loss while being valid under symmetry, and further use a Generative Adversarial Network to lock on to a plausible mode.
In addition to dense correspondences, SO-Pose~\cite{sopose} also estimates self-occlusion maps to constrain their pose hypotheses.
CDPN~\cite{cdpn} regresses dense correspondences but only use them to estimate the rotation, 
and instead uses a separate head to regress the translation.
DPODv2~\cite{dpodv2} also regresses dense correspondences. They obtain an initial pose with PnP-RANSAC and refine the pose based on differentiable rendering and the estimated dense correspondences. They handle symmetry similar to BB8 and PVNet.
\cite{local-surface-embeddings} regresses dense hand-crafted 3D features
to establish 2D-3D correspondences, 
even for objects not seen during training.
While their feature definition handles symmetries implicitly, 
they are by definition limited to local, rotation invariant 3D information.
%
EPOS~\cite{epos} learns a dense 2D-3D correspondence distribution and handles symmetries implicitly. 
They do so by discretizing the surface into a set of fragments and predicting a probability distribution over fragments per pixel.
They regress a coordinate offset per fragment to increase correspondence accuracy.

EPOS is the most related pose estimation method, being the only one representing a distribution of correspondences.
In contrast to all pose estimation methods, we estimate a \textit{continuous} distribution over the object surface, 
and we use the distributions instead of correspondences for pose scoring and refinement.

There are also works outside object pose estimation, that aim to establish correspondences.
\cite{dense-object-nets} learns embeddings in a self-supervised fashion with a contrastive loss
to establish 2D-2D correspondences across images for object manipulation.
\cite{continuous} proposes to establish dense 2D-3D correspondences using continuous surface embeddings, and their framework for learning correspondences is similar to ours with some important differences.
Most importantly, they only establish a single correspondence per pixel and do not discuss or show the ability to represent distributions or in other ways handle ambiguities.
Nor do they use the obtained correspondences for object pose estimation.
There are more differences, including that
their key model embeddings are explicitly regularized to be continuous, 
whereas the continuity of our embeddings are primarily driven by visual ambiguity, 
and that they use a fixed set of object vertices during training, 
whereas we use a contrastive loss, sampling from the surface during training.

%% file: sections/method.tex
\section{Methods}
This section presents our pose estimation method. 
We begin with a short overview,
then describe in more detail how we represent and learn correspondence distributions 
and lastly how pose estimates are obtained from those distributions.

\subsection{Overview}
Similar to other approaches \cite{pix2pose, cosypose, sopose, deepim, cdpn}, we base our method on image crops from a detection model.
We feed an image crop through our model to obtain dense (pixel-wise) surface distributions and a mask which together form a correspondence distribution.
In a PnP-RANSAC fashion, we sample pose hypotheses from the correspondence distribution and score each hypothesis based on the mask and surface distributions.
The best scoring pose hypothesis is then refined based on the surface distributions to obtain the final pose estimate.

\subsection{Learning Correspondence Distributions}
Given an image crop, $I \in \R^{H \times W \times 3}$, of an object 
and image-coordinates, $u$, inside the mask of the object, $u \in M$, 
we aim to learn a surface distribution describing which point, $c$, 
on the object surface, $c \in S \subset \R^3$, the pixel corresponds to
\begin{equation}
    p(c | I, u, u \in M).
\label{eq:prob-dense}
\end{equation}
We consider the mask to be the set of pixels, where the object is present, even if it is occluded.

Let $q \in \R^E$ be a query and $k \in \R^E$ be a key in a shared embedding space with dimension $E$, and let each surface point, $c_i$, be represented by a corresponding key, $k_i$.
Similar to \cite{continuous}, given a query, we define the surface distribution over a discrete set of surface points, $\tilde S \subset S$, as the softmax over dot products between the query and all keys 
\begin{equation}
    \Pr(c_i | q, \tilde S) = \frac{
        \exp(q^T k_i)
    }{
        \sum_{c_j \in \tilde S} \exp(q^T k_j)
    }, \quad c_i \in \tilde S.
\label{eq:cond-corr-prob}
\end{equation}

We use a \textit{key model}, a small fully connected network, $g: \R^3 \mapsto \R^E$, to map surface points to keys, and a \textit{query model}, an encoder-decoder convolutional network, $f: \R^{H \times W \times 3} \mapsto \R^{H \times W \times E}$, to map color images to query images.
With the set of surface points, $\tilde S$, sampled uniformly from the object surface, and a set of image coordinates, $\tilde U = \{u_1, \ldots, u_N\}$, sampled uniformly from the object mask,
we define the embedding loss as the InfoNCE~\cite{info-nce} loss
\begin{equation}
\label{eq:corr-loss}
    L_\text{E} = - 
    \frac{1}{|\tilde U|}
    \sum_{u \in \tilde U} \log \dfrac{
        \exp(q_u^T k_u)
    }{
        \sum_{c_i \in \tilde S \cup c_u} \exp(q_u^T k_i)
    },
\end{equation}
where $q_u$ refers to the query in the query image at $u$, and $k_u = g(c_u)$ refers to the key from the object coordinate, $c_u$, that is present at $u$ (Fig. \ref{fig:training_overview}).
\cite{info-nce} shows that optimizing this loss will result in estimating the probability density ratio 
\begin{equation}
    \exp(q^T k_i) \propto \frac{p(c_i|q)}{p(c_i)},
\end{equation}
regardless of the sample size, $|\tilde S|$.
Since we sample $c_i$ uniformly from the surface, 
it follows that $\exp(q^T k_i) \propto p(c_i|q)$, which means that we by normalizing over the object surface
are indeed estimating the probability density function in Eq.~\ref{eq:prob-dense}
\begin{equation}
    p(c_i | I, u, u \in M) = \frac{\exp(q_u^T k_i)}{\oiint_{c_j \in S} \exp(q_u^T k_j)}.
\end{equation}

Until now, we have only looked at surface distributions, 
assuming pixel coordinates to be inside the object mask.
To establish a distribution over correspondences, $p(c, u|I)$,
we add a channel in the same convolutional architecture as the query model
to estimate the object mask, and let the discrete distribution over image coordinates, $\Pr(u|I)$, 
be proportional to the estimated probability that the object is present at pixel $u$
\begin{equation}
    \Pr(u|I) \propto \Pr(u \in M | I, u).
\label{eq:u-dist}
\end{equation}
The distribution over correspondences is then modeled by
\begin{equation}
    p(c, u | I) = \Pr(u | I)p(c | I, u).
\label{eq:corr-prob}
\end{equation}

With $L_M$ being the average binary cross-entropy loss for the mask, the total loss is
\begin{equation}
    L = L_E + L_M.
\end{equation}

\paragraph{Notes on Representation.}
While some simple visual ambiguities, like global symmetry, can be modelled explicitly, 
it is questionable whether it is feasible to explicitly model all relevant ambiguities.
Learning ambiguities from data requires the model to be able to represent the ambiguities, 
and while the representation could in theory be in pose space, we are not aware of any work 
successfully learning multi-modal object pose distributions directly.

The dense query image can be seen as a proxy for a pose distribution.
The key model represents an object-specific surface distribution model with $E$ parameters, where the query model estimates the model parameters pixel-wise.
Maximizing the probability of the ground truth coordinates (Eq. \ref{eq:corr-loss}) encourages the key model to model the types of visual ambiguities that are most common in the data, and only those, leading to an efficient representation.
In case of global symmetry, the key model can simply learn to map symmetric points to the same keys, but it can also represent more complex and view-dependent ambiguities.

To provide intuition on the representational capacity of the embeddings,
let us imagine mapping an object's surface onto the surface of a 3D sphere and let those 3D coordinates represent keys. 
We then see that a query identifies any key uniquely when it has the same direction as the key and an infinite norm.
A lower query norm gradually leaks more and more distribution mass onto neighbouring keys.
Three embedding dimensions are thus enough to represent distributions with simple ambiguities,
but more embedding dimensions allow the models to represent more ambiguities.



\subsection{From Embeddings to Pose}
At inference, 2D-3D correspondences are established based on the estimated correspondence distribution from Equation \ref{eq:corr-prob}. 
The correspondences form pose hypotheses which are scored based on the agreement with the estimated mask and correspondence distributions,
and the best scoring pose hypothesis is refined by a local maximization of the probability of the visible surface coordinates.

\paragraph{Sampling Pose Hypotheses.}
While the 2D-3D correspondence distribution is continuous in the sense that it can be evaluated for any point on the surface, we choose a large set of points, $\hat S \subset S, |\hat S| \approx 75.000$, sampled evenly from the object surface, using \cite{even-sampling} as implemented in Meshlab\footnote{https://www.meshlab.net/}.

For each query, the probability distribution over keys is computed and multiplied with the mask probability to obtain the correspondence probabilities (Eq.~\ref{eq:corr-prob}), which can be sampled from efficiently with inversion sampling.
Sampling directly from this distribution results in an even sampling inside the estimated mask.
To sample image coordinates with lower entropy more often,
we sharpen the correspondence distribution by sampling proportional to $\Pr(c,u|I)^\gamma$ instead, where $\gamma = 1.5$ is chosen qualitatively.
Pose hypotheses are found from correspondences by AP3P~\cite{ap3p} as implemented in OpenCV\footnote{https://opencv.org/}.

Since we are trying to establish correspondences only for the part of the surface that is not self-occluded, 
the correspondences used to obtain a pose hypothesis should be visible under the pose hypothesis.
We discard the pose hypotheses where this is not the case based on the correspondence normals.

\paragraph{Pose Hypothesis Scoring.}
We propose a pose score based on the training loss, which has two parts.
The first part represents the agreement with the estimated mask, and
the second part represents the agreement with the estimated correspondence distributions.
We define the mask score as the estimated average log likelihood of the pose hypothesis mask
\begin{equation}
    s_M = \frac{1}{|U|} \sum_{u\in U} \log \Pr(\hat M_u),
\end{equation}
where $U$ is the discrete set of all pixel coordinates in the image, and $\hat M_u$ denotes whether or not the object is present at $u$ under the pose hypothesis. 
We define the correspondence score as the estimated average log likelihood of the visible surface coordinates of the pose hypothesis
\begin{equation}
    s_C = \frac{1}{|\hat M|} \sum_{u\in \hat M} \log \Pr(c_u|q_u, u \in M).
\label{eq:corr-score}
\end{equation}
To relate the two scores, we normalize by their maximum entropy
\begin{equation}
    s = \frac{s_M}{\log(2)} + \frac{s_C}{\log(|\hat S|)}.
\end{equation}

$\Pr(c|q)$ is calculated once for all query-key pairs, obtaining a table, $l \in R^{H\times W\times |\hat S|}$, used for the inversion sampling,
and for efficient computation of Eq.~\ref{eq:corr-score} by indexing.
The table has memory requirements proportional to $|U| \times |\hat S|$, and we use square images of resolution 224, so we downscale the query image by a factor of three to make it fit in GPU memory.

For the correspondence score, we max pool $l$ spatially with a kernel size three and stride one. 
Max pooling is intended to make the score more robust, as pose hypotheses will not be penalized for being a few pixels off. 
This is analog to the reprojection error threshold in common PnP-RANSAC~\cite{ransac} frameworks.
Max pooling does compromise precision in favor of robustness, but ideally refinement makes up for this compromise. 

Rasterization is relatively expensive, so instead, we project all object coordinates in $\hat S$ into the image and choose only the object coordinate closest to the camera per pixel.

\paragraph{Refinement.}
The pose hypothesis with the best score is refined by local maximization of the correspondence score in Equation \ref{eq:corr-score}.
We start by rendering the visible object coordinates under the initial pose hypothesis.
For the optimization objective, we then project the object coordinates into the image under the current pose to sample the query image with bilinear interpolation and evaluate the numerator in Equation \ref{eq:cond-corr-prob}.
As evaluating the denominator is computationally expensive, it is instead pre-computed for all queries initially to obtain a denominator-image which is also sampled by bilinear interpolation.

The objective is maximized with BFGS as implemented in SciPy\footnote{https://scipy.org/}, and the gradient is computed by automatic differentiation in PyTorch\footnote{https://pytorch.org/}.

\paragraph{Using Depth Images.}
Estimating accurate depth from a single color image is notoriously hard.
When a depth image is available, we adjust the depth of our pose estimate.

We find the set of image coordinates where the query norm is at least 80\% of the maximum query norm across the image.
The high query norm indicates that the model is certain, so we assume these coordinates to correspond to visible parts of the target object.
We find the difference between the depth image and estimated depth under the pose hypothesis for each of those coordinates
and choose the depth adjustment as the median of those differences.

We need to choose a ray to adjust the depth along.
We use the ray that goes through the center of mass of the chosen image coordinates.

\paragraph{Architecture and Training Details.}
For the query model, we use a U-Net~\cite{unet} architecture with an ImageNet~\cite{imagenet} pretrained ResNet-18~\cite{resnet} backbone.
As MLPs with ReLU activations have shown to be biased towards low frequencies in their input~\cite{low-freq-bias}, we use the Siren~\cite{siren} MLP for the key model with the motivation that locally consistent embeddings should be a result of visual ambiguity, not limitations of the MLP.

We make heavy use of image augmentations during training. 
We add translation and scale noise to the ground truth crops to simulate detector inaccuracies and sample image rotations uniformly.
From the Albumentations\footnote{https://albumentations.ai/} library, we use GaussianBlur, ISONoise, GaussNoise, CLAHE, CoarseDropout and ColorJitter. 
We further add debayering artifacts and unsharp masking.
All image augmentations have a 50 \% chance of being applied.

We implement and train our models in Python with PyTorch.
We use Adam~\cite{adam} with a learning rate of $3\cdot 10^{-4}$ for the query model and $3\cdot 10^{-5}$ for the key model, warming up the learning rate linearly over the first 2000 steps.
We use a batch size of 16 and sample 1024 query-key pairs inside the object mask and 1024 negative keys from uniform surface samples per crop (Fig. \ref{fig:training_overview}).
We use 12 embedding dimensions and have a separate key model per object and a separate decoder per object in the query model. The query encoder is shared across objects in a dataset.
We train and run our models on an Nvidia RTX 2080. 

%% file: sections/experiments.tex
\section{Experiments}

\begin{table*}
    \centering
    \small
    \begin{tabular}{llc|lllllll|l}
        Method & Domain & Synth
        & LM-O & T-LESS & TUD-L & IC-BIN & ITODD & HB & YCB-V & Avg  \\ \toprule
        \textbf{SurfEmb (ours)} & RGB & \cmark & \textbf{0.656} & \textbf{0.741} & \textbf{0.715} & \textbf{0.585} & \textbf{0.387} & \textbf{0.793} & \textbf{0.653} & \textbf{0.647} \\
        Epos \cite{epos} & RGB & \cmark & 0.547 & 0.467 & 0.558 & 0.363 & 0.186 & 0.580 & 0.499 & 0.457 \\
        CDPNv2 \cite{cdpn} & RGB & \cmark & 0.624 & 0.407 & 0.588 & 0.473 & 0.102 & 0.722 & 0.390 & 0.472 \\
        DPODv2 \cite{dpodv2} & RGB & \cmark & 0.584 & 0.636 & - & - & - & 0.725 & - & - \\ 
        PVNet \cite{pvnet} & RGB & \cmark & 0.575 & - & - &- &- &-&-&- \\
        CosyPose \cite{cosypose} & RGB & \cmark & 0.633 & 0.640 & 0.685 & 0.583 & 0.216 & 0.656 & 0.574 & 0.570 \\ 
        \midrule
        \textbf{SurfEmb (ours)} & RGB & \xmark & \textbf{0.656} & \textbf{0.770} & 0.805 & \textbf{0.585} & \textbf{0.387} & \textbf{0.793} & 0.718 & \textbf{0.673} \\
        Pix2Pose \cite{pix2pose} & RGB & \xmark & 0.363 & 0.344 & 0.420 & 0.226 & 0.134 & 0.446 & 0.457 & 0.342 \\
        CDPNv2 \cite{cdpn} & RGB & \xmark & 0.624&0.478&0.772&0.473&0.102&0.722&0.532&0.529 \\
        SO-Pose \cite{sopose} & RGB & \xmark & 0.613 & - & - & - & - & - & 0.715 & - \\
        CosyPose \cite{cosypose} & RGB & \xmark & 0.633&0.728&\textbf{0.823}&0.583&0.216&0.656&\textbf{0.821} & 0.637 \\ 
        \midrule
        \textbf{SurfEmb (ours)} & RGB-D & \cmark & \textbf{0.758} & \textbf{0.828} & 0.854 & \textbf{0.656} & 0.498 & \textbf{0.867} & 0.806 & \textbf{0.752} \\
        \textbf{SurfEmb (ours)} & RGB-D & \xmark & \textbf{0.758} & \textbf{0.833} & 0.933 & \textbf{0.656} & 0.498 & \textbf{0.867} & 0.824 & \textbf{0.767} \\
        Drost \cite{drost} & RGB-D & * & 0.515&0.500&0.851&0.368&\textbf{0.570}&0.671&0.375 & 0.550 \\ 
        Vidal Sensors \cite{vidal} & D & * & 0.582 & 0.538 & 0.876 & 0.393 & 0.435 & 0.706 & 0.450 & 0.569 \\ 
        Koenig-Hybrid \cite{konig} & RGB-D & \xmark & 0.631&0.655&0.920&0.430&0.483&0.651&0.701 & 0.639 \\
        Pix2Pose \cite{pix2pose} & RGB-D & \xmark & 0.588&0.512&0.820&0.390&0.351&0.695&0.780&0.591 \\
        CosyPose \cite{cosypose} & RGB-D & \xmark & 0.714 & 0.701 & \textbf{0.939} & 0.647 & 0.313 & 0.712 & \textbf{0.861} & 0.698 \\ 
        \bottomrule 
    \end{tabular}
    \medskip
    \caption{
        Average Recall on the BOP Core datasets. 
        Methods are split into three groups; RGB methods trained purely on synthetic data, RGB methods that are also trained on real data for T-LESS, TUD-L and YCB-V, and methods using depth.
        The best performing method for each dataset in each group is marked in bold.
        Synth: Trained purely on the synthetic data provided by the BOP Challenge. 
        Note that real training data is only available for three of the datasets, so some entries are duplicates.
        * Methods do not use the provided training images.
    }
\label{tab:main}
\end{table*}

We evaluate our method on the BOP Challenge~\cite{bop-challenge}, which is not yet saturated and arguably the most comprehensive rigid body pose estimation benchmark with seven datasets; LM-O~\cite{lmo}, T-LESS~\cite{tless}, TUD-L~\cite{bop-challenge}, IC-BIN~\cite{icbin}, ITODD~\cite{itodd}, HB~\cite{hb} and YCB-V~\cite{posecnn}.
The task involves pose estimation from a single image with a varying number of instances of varying number of objects (ViVo).
BOP evaluates three pose errors based on visible surface discrepancy (VSD), maximum symmetry-aware surface distance (MSSD), and maximum symmetry-aware projection distance (MSPD). 
An average recall is computed for each of the errors, AR\textsubscript{VSD}, AR\textsubscript{MSSD}, AR\textsubscript{MSPD},
based on a set of error thresholds, and \textit{the} average recall, AR, refers to the average of the three average recalls.

We train our models on the synthetic, physically based rendering datasets provided by the BOP Challenge.
Because the datasets vary in number of objects and instances, we train our models 5 epochs for IC-BIN, 10 epochs for T-LESS, ITODD, HB and YCBV and 20 epochs for LM-O and TUD-L.
Training time is 1-2 days on a single Nvidia RTX 2080, 
which is significantly less compute than \eg CosyPose~\cite{cosypose} 
that trains on 32 Nvidia V100 for 10 hours.

We also fine-tune on the real training datasets which are provided for T-LESS, TUD-L and YCB-V.
Since the real dataset for T-LESS only contains separated objects with a black background, we fine-tune on a mixture of synthetic and real images for this dataset.

During inference, we use the available detection crops from the models in CosyPose~\cite{cosypose} trained on the synthetic images.
We use a simple test-time augmentation, where the input is rotated 0, 90, 180 and 270 degrees, fed through the model, rotated back and averaged.

Inference time is approximately 2.2s per image crop, 
including around 20ms for the query model forward pass, 
1.2s for PnP-RANSAC,
and 1.0s for pose refinement.
Our implementation is written in Python.
A compiled version could reduce inference time.
We also use a fixed number of 20.000 iterations during PnP-RANSAC.
Early termination and further pose pruning could reduce the PnP-RANSAC time.

\begin{figure}
    \centering
    \includegraphics[width=1\linewidth]{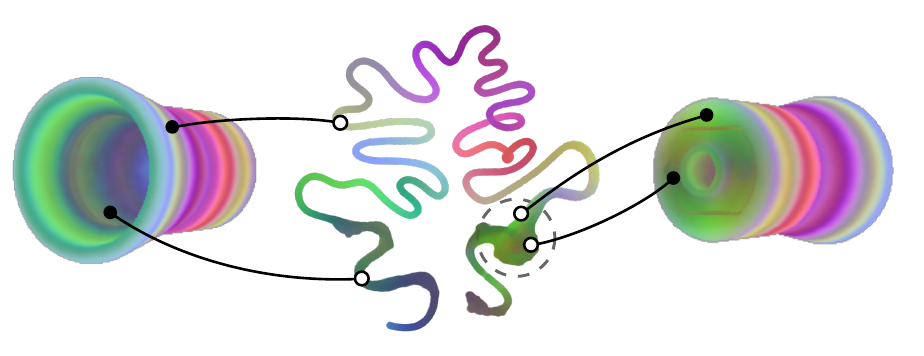}
    \caption{
        2D-projection with UMAP of learnt key embeddings.
        The key model has learnt the expected one-dimensional manifold 
        for an object with continuous rotational symmetry.
        Note that the manifold dimensionality expands where the symmetry breaks as indicated by the dashed circle.
        The global embedding structure of most other objects are difficult to convey in 2D, 
        but they are available in 3D on the project site.
    }
    \label{fig:manifold}
\end{figure}

\paragraph{Main Results.}
Our main results are presented in Tab.~\ref{tab:main}.
For ITODD, we show a 79 \% relative improvement over the next best RGB method.
Our RGB method trained on synthetic data is state-of-the-art on T-LESS and HB, even compared with methods trained on real data and methods using depth.

Our method is state-of-the-art across all datasets compared with other methods trained on synthetic data.
Our RGBD method trained purely on synthetic data is state-of-the-art on the BOP Challenge, even compared to methods that train on real data.

\paragraph{Visualizing Embeddings.}
We reduce embeddings to three dimensions for visualization by summation over each of three distinct sets of embedding dimensions, and normalize the range into the valid RGB range, making sure to have zero visualized by gray.
Key embeddings are further demeaned before summation.
Summation over embedding dimensions does mean ambiguity in the visualization, but we find that it better portrays the embedding manifold in practice than subsampling dimensions or PCA.

\begin{figure}
    \centering
    \includegraphics[width=1\linewidth]{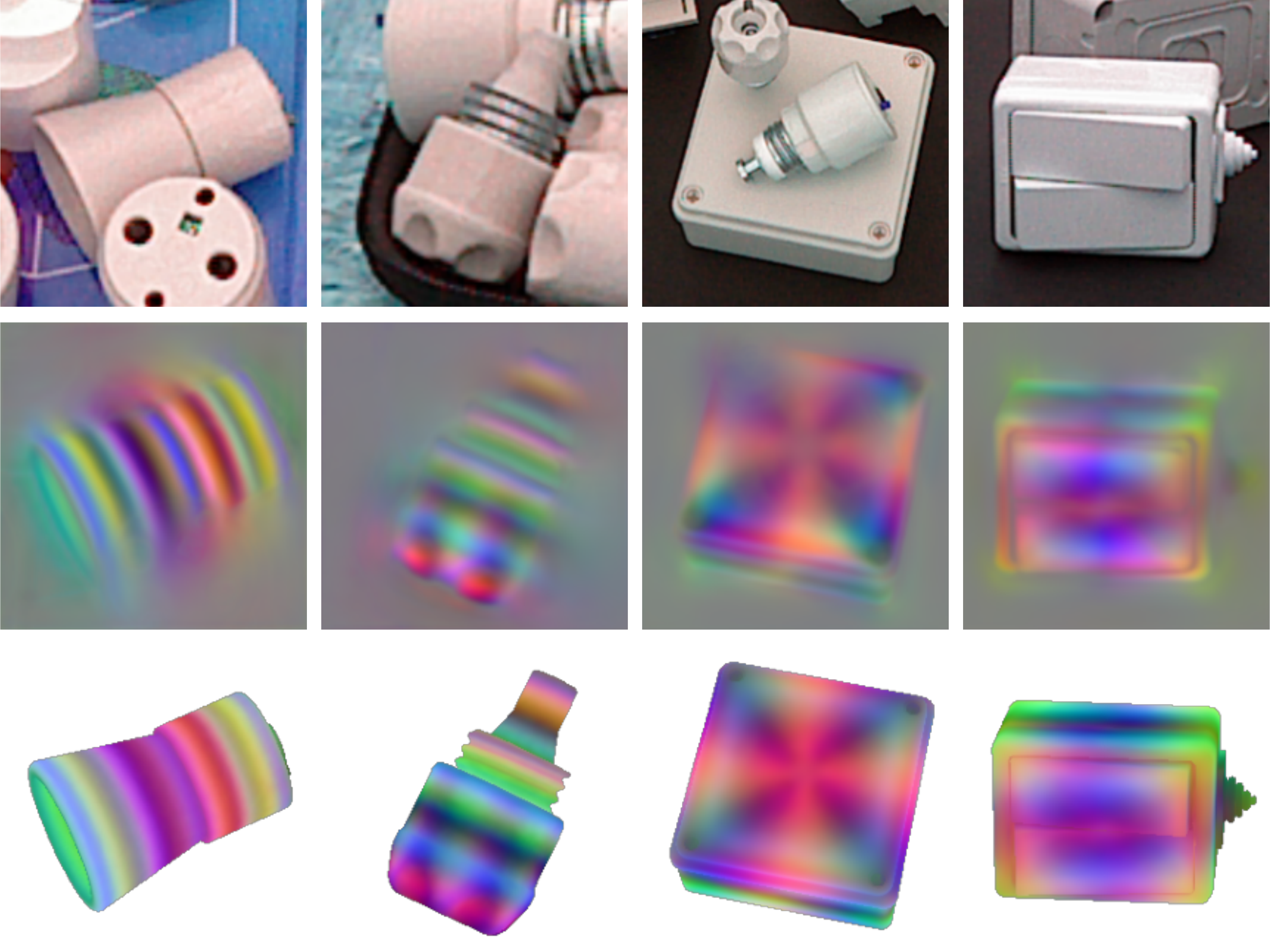}
    \caption{
        Embedding examples. 
        From top to bottom; Input image, query image, keys under the ground truth pose.
        While the query model has only been trained within the full mask of the objects, it has learnt to output low-norm queries (close to gray) outside the mask, indicating high entropy distributions.
    }
    \label{fig:emb_examples}
\end{figure}

\paragraph{Qualitative Results.}
Our models learn embeddings (Fig.~\ref{fig:emb_examples}) that are meaningful in the sense that surface points 
that are visually similar, \eg because of symmetry, have similar embeddings.
We find the key embeddings for objects especially interesting as they represent the 
learnt latent models of the objects, and we find that those latent models align well with our own mental models.
For an object with continuous rotational symmetry, we inspect the key embedding manifold (Fig.~\ref{fig:manifold}) using UMAP~\cite{umap} and show that the key model has learnt the expected one-dimensional embedding manifold without any prior knowledge about the symmetry.

\begin{figure}
    \centering
    \includegraphics[width=1\linewidth]{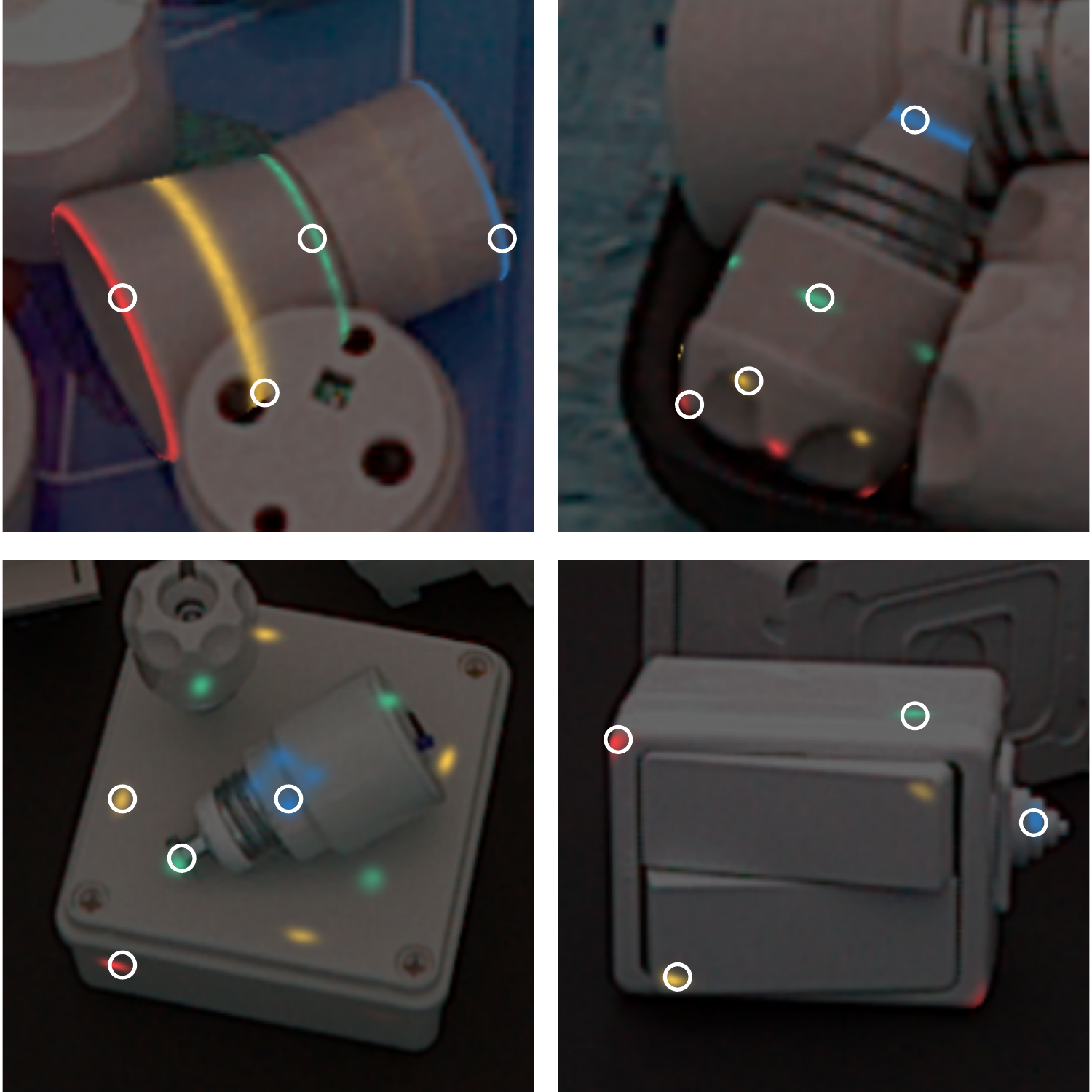}
    \caption{
        Correspondence distribution examples on T-LESS objects with varying degrees of symmetry.
        Each image is fed through the query model to obtain a query image from which four queries are chosen.
        For each query, the distribution over keys from the object surface are imposed on the image in the ground truth pose.
        The positions of the queries are marked with circles, and each distribution is shown in a distinct color. 
        For clarity, distribution mass is only shown for the part of the surface that is not self-occluded.
        Interactive examples are available on the project site.
    }
    \label{fig:dist_examples}
\end{figure}

We also show that the models trained on synthetic data produce accurate correspondence distributions on real images (Fig.~\ref{fig:dist_examples}).
Note that distributions from occluded queries (top left, yellow) have higher entropy distributions, 
and in the case where there is no visual ambiguity (bottom right, yellow and red),
the way that the distribution is imprecise is still meaningful, giving most probability to the correct modes and some probability to visually similar modes.
To the best of our knowledge, no other work has shown correspondence distributions of this quality.

\begin{figure*}
    \centering
    \includegraphics[width=1\linewidth]{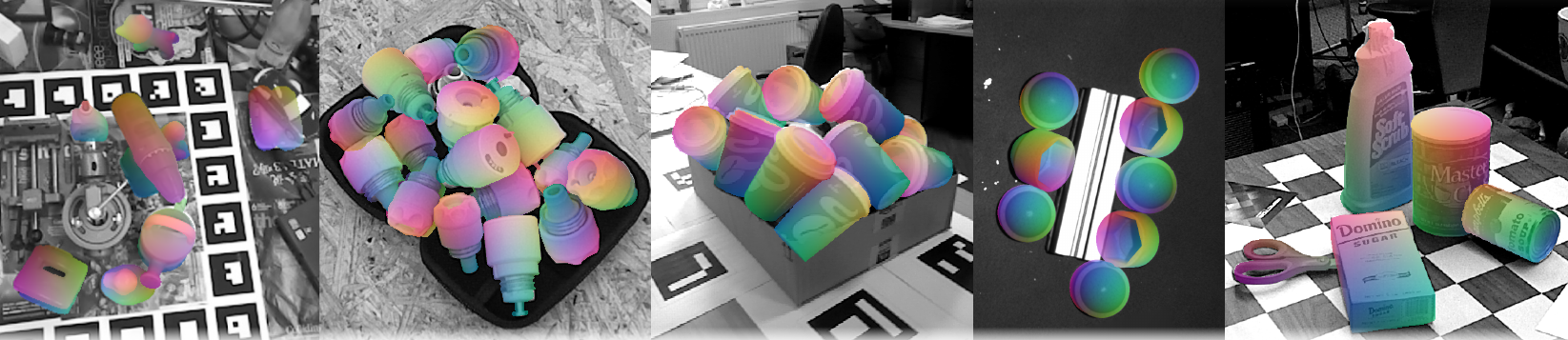}
    \caption{
        Pose estimation examples on different datasets. 
        From left to right; LM-O, T-LESS, IC-BIN, ITODD and YCB-V.
    }
    \label{fig:pose_examples}
\end{figure*}

Pose estimation examples are shown in Fig.~\ref{fig:pose_examples}.

\paragraph{Ablation Study.}
We investigate the effect of different hyper-parameters in Tab.~\ref{tab:ablation}.
Specifically, we look at 
six vs twelve embedding dimensions, 
a single decoder (but with object-specific last layers) vs a separate decoder per object, 
with or without refinement, 
and with or without test time augmentation.

More embedding dimensions, separate decoders and test time augmentation all show steady improvements,
and refinement shows a significant improvement.
The forward pass on the query model is computationally inexpensive compared to pose scoring and pose refinement,
so test time augmentation is an effective way to improve performance.
We hypothesize that test time augmentation improves performance by 
providing a smoother query image, leading to a better posed refinement optimization problem.


\begin{table}
    \centering
    \small
    \resizebox{\linewidth}{!}{%
        \begin{tabular}{llll|lll|l}
            E & SD & R & TA & AR\textsubscript{VSD} & AR\textsubscript{MSSD} & AR\textsubscript{MSPD} & AR \\ \toprule
            6 & \xmark & \xmark & \xmark & 0.434 & 0.465 & 0.813 & 0.571 \\
            6 & \xmark & \cmark & \xmark & 0.616 & 0.642 & 0.831 & 0.696 \\
            12 & \xmark & \xmark & \xmark & 0.466 & 0.497 & 0.822 & 0.595 \\
            12 & \xmark & \cmark & \xmark & 0.643 & 0.669 & 0.843 & 0.719 \\
            12 & \cmark & \xmark & \xmark & 0.502 & 0.536 & 0.835 & 0.624 \\
            12 & \cmark & \cmark & \xmark & 0.653 & 0.682 & 0.851 & 0.729 \\
            12 & \cmark & \cmark & \cmark & 0.668 & 0.696 & 0.860 & 0.741 \\
            \bottomrule 
        \end{tabular}%
    }
    \medskip
    \caption{
        Ablation study of our method on T-LESS~\cite{tless}. 
        E: number of Embedding dimensions,
        SD: a Separate Decoder per object,
        R: with Refinement.
        TA: Test time rotation Augmentation.
    }
\label{tab:ablation}
\end{table}

\paragraph{Failure Cases.}
Not surprisingly, we observe that our method trained on synthetic data performs best, 
when the CAD models represent the real objects well.
For some objects, like the bowl and cup in YCB-V, the CAD models contain baked-in reflections. 
Since these reflections are always present in the data, even under heavy data augmentation, our models learn to rely on these.

For other objects, like the models in TUD-L and IC-BIN and some objects in LM-O and ITODD, the CAD model geometry appear to be significantly different from the real surfaces, 
which leads to sub-optimal pose scoring and refinement on the real images.

\begin{figure}
    \centering
    \includegraphics[width=1\linewidth]{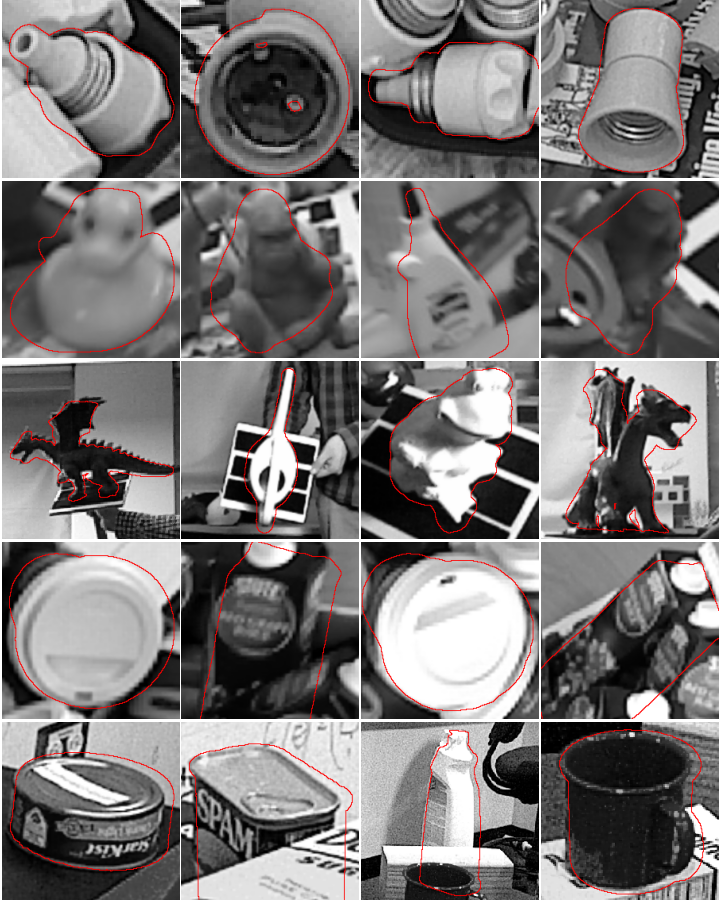}
    \caption{
        Poor ground truth poses - \textbf{not} pose estimates.
        The rows correspond to 
        T-LESS, LM-O, TUD-L, IC-BIN and YCB-V, respectively. 
        For each dataset, we choose the qualitatively poorest ground truth poses looking at 50 poses per dataset.
        The outline of the objects in their ground truth poses are shown in red.
    }
    \label{fig:gt_poses}
\end{figure}

We find that some ground truth poses are surprisingly poor.
This is especially apparent for LM-O, TUD-L, IC-BIN and YCB-V, as seen in Fig. \ref{fig:gt_poses}.
Ground truth poses for HB and ITODD are not publicly available.
Poor ground truth poses on the test datasets set an upper bound on the achievable performance,
unless the pose errors are biased and can be learnt from the training data.
In that case, methods that replicate this bias obtain optimistic results, questioning whether low recall thresholds should be used for such datasets.

Poor ground truth poses and model geometry together with our method's focus on object surfaces could explain why our method does not seem to improve as much as \eg CosyPose~\cite{cosypose} when fine-tuning on real data on TUD-L and YCB-V.

%% file: sections/limitations.tex
\section{Future Work and Limitations}
We have trained our models with limited compute, we have only done limited parameter search, and we use a U-Net~\cite{unet} architecture for the query model.
More compute, other hyper parameters and other encoder-decoder architectures may improve performance.
Our method only leverages information from a single-view but pose sampling, pose evaluation and pose refinement could all be extended to multi-view variants.
Our refinement method assumes that the visible object surface does not change significantly during refinement,
and we have observed refinement failing because of this assumption.
Our method is effectively a four-stage approach: detections from an image, distributions from a crop, initial pose from distributions and refinement. 
This is similar to related methods but indisputably makes our pipeline more complex than end-to-end approaches.

%% file: sections/conclusion.tex
\section{Conclusion}
We have proposed a way to learn dense and continuous 2D-3D correspondence distributions from color images.
The distributions are learnt implicitly from data with a contrastive loss and no prior knowledge about ambiguities like symmetry.
The distributions are represented in object-specific latent spaces by
an encoder-decoder query model and a small fully connected key model.
We have also proposed a new pose estimation method using these distributions for
pose sampling, pose scoring and pose refinement.
Our method is unsupervised with respect to visual ambiguities, 
yet we have shown that the query- and key models learn to represent accurate multi-modal surface distributions.
Our pose estimation method has improved the state-of-the-art significantly on the comprehensive BOP Challenge,
trained purely on synthetic data, even compared with methods trained on real data.


%% file: main.bbl
\begin{thebibliography}{10}\itemsep=-1pt

\bibitem{lmo}
Eric Brachmann, Alexander Krull, Frank Michel, Stefan Gumhold, Jamie Shotton,
  and Carsten Rother.
\newblock Learning 6d object pose estimation using 3d object coordinates.
\newblock In {\em European conference on computer vision}, pages 536--551.
  Springer, 2014.

\bibitem{even-sampling}
Massimiliano Corsini, Paolo Cignoni, and Roberto Scopigno.
\newblock Efficient and flexible sampling with blue noise properties of
  triangular meshes.
\newblock {\em IEEE transactions on visualization and computer graphics},
  18(6):914--924, 2012.

\bibitem{imagenet}
Jia Deng, Wei Dong, Richard Socher, Li-Jia Li, Kai Li, and Li Fei-Fei.
\newblock Imagenet: A large-scale hierarchical image database.
\newblock In {\em 2009 IEEE conference on computer vision and pattern
  recognition}, pages 248--255. Ieee, 2009.

\bibitem{poserbpf}
Xinke Deng, Arsalan Mousavian, Yu Xiang, Fei Xia, Timothy Bretl, and Dieter
  Fox.
\newblock Poserbpf: A rao--blackwellized particle filter for 6-d object pose
  tracking.
\newblock {\em IEEE Transactions on Robotics}, 37(5):1328--1342, 2021.

\bibitem{sopose}
Yan Di, Fabian Manhardt, Gu Wang, Xiangyang Ji, Nassir Navab, and Federico
  Tombari.
\newblock So-pose: Exploiting self-occlusion for direct 6d pose estimation.
\newblock In {\em Proceedings of the IEEE/CVF International Conference on
  Computer Vision}, pages 12396--12405, 2021.

\bibitem{icbin}
Andreas Doumanoglou, Rigas Kouskouridas, Sotiris Malassiotis, and Tae-Kyun Kim.
\newblock Recovering 6d object pose and predicting next-best-view in the crowd.
\newblock In {\em Proceedings of the IEEE conference on computer vision and
  pattern recognition}, pages 3583--3592, 2016.

\bibitem{itodd}
Bertram Drost, Markus Ulrich, Paul Bergmann, Philipp Hartinger, and Carsten
  Steger.
\newblock Introducing mvtec itodd-a dataset for 3d object recognition in
  industry.
\newblock In {\em Proceedings of the IEEE International Conference on Computer
  Vision Workshops}, pages 2200--2208, 2017.

\bibitem{drost}
Bertram Drost, Markus Ulrich, Nassir Navab, and Slobodan Ilic.
\newblock Model globally, match locally: Efficient and robust 3d object
  recognition.
\newblock In {\em 2010 IEEE computer society conference on computer vision and
  pattern recognition}, pages 998--1005. Ieee, 2010.

\bibitem{ransac}
Martin~A Fischler and Robert~C Bolles.
\newblock Random sample consensus: a paradigm for model fitting with
  applications to image analysis and automated cartography.
\newblock {\em Communications of the ACM}, 24(6):381--395, 1981.

\bibitem{dense-object-nets}
Peter~R Florence, Lucas Manuelli, and Russ Tedrake.
\newblock Dense object nets: Learning dense visual object descriptors by and
  for robotic manipulation.
\newblock {\em arXiv preprint arXiv:1806.08756}, 2018.

\bibitem{resnet}
Kaiming He, Xiangyu Zhang, Shaoqing Ren, and Jian Sun.
\newblock Deep residual learning for image recognition.
\newblock In {\em Proceedings of the IEEE conference on computer vision and
  pattern recognition}, pages 770--778, 2016.

\bibitem{epos}
Tomas Hodan, Daniel Barath, and Jiri Matas.
\newblock Epos: Estimating 6d pose of objects with symmetries.
\newblock In {\em Proceedings of the IEEE/CVF conference on computer vision and
  pattern recognition}, pages 11703--11712, 2020.

\bibitem{tless}
Tom{\'a}{\v{s}} Hodan, Pavel Haluza, {\v{S}}tep{\'a}n Obdr{\v{z}}{\'a}lek, Jiri
  Matas, Manolis Lourakis, and Xenophon Zabulis.
\newblock T-less: An rgb-d dataset for 6d pose estimation of texture-less
  objects.
\newblock In {\em 2017 IEEE Winter Conference on Applications of Computer
  Vision (WACV)}, pages 880--888. IEEE, 2017.

\bibitem{bop-challenge}
Tom{\'a}{\v{s}} Hoda{\v{n}}, Martin Sundermeyer, Bertram Drost, Yann Labb{\'e},
  Eric Brachmann, Frank Michel, Carsten Rother, and Ji{\v{r}}{\'\i} Matas.
\newblock Bop challenge 2020 on 6d object localization.
\newblock In {\em European Conference on Computer Vision}, pages 577--594.
  Springer, 2020.

\bibitem{hb}
Roman Kaskman, Sergey Zakharov, Ivan Shugurov, and Slobodan Ilic.
\newblock Homebreweddb: Rgb-d dataset for 6d pose estimation of 3d objects.
\newblock In {\em Proceedings of the IEEE/CVF International Conference on
  Computer Vision Workshops}, pages 0--0, 2019.

\bibitem{ap3p}
Tong Ke and Stergios~I Roumeliotis.
\newblock An efficient algebraic solution to the perspective-three-point
  problem.
\newblock In {\em Proceedings of the IEEE Conference on Computer Vision and
  Pattern Recognition}, pages 7225--7233, 2017.

\bibitem{adam}
Diederik~P Kingma and Jimmy Ba.
\newblock Adam: A method for stochastic optimization.
\newblock {\em arXiv preprint arXiv:1412.6980}, 2014.

\bibitem{konig}
Rebecca K{\"o}nig and Bertram Drost.
\newblock A hybrid approach for 6dof pose estimation.
\newblock In {\em European Conference on Computer Vision}, pages 700--706.
  Springer, 2020.

\bibitem{cosypose}
Yann Labb{\'e}, Justin Carpentier, Mathieu Aubry, and Josef Sivic.
\newblock Cosypose: Consistent multi-view multi-object 6d pose estimation.
\newblock In {\em European Conference on Computer Vision}, pages 574--591.
  Springer, 2020.

\bibitem{deepim}
Yi Li, Gu Wang, Xiangyang Ji, Yu Xiang, and Dieter Fox.
\newblock Deepim: Deep iterative matching for 6d pose estimation.
\newblock In {\em Proceedings of the European Conference on Computer Vision
  (ECCV)}, pages 683--698, 2018.

\bibitem{cdpn}
Zhigang Li, Gu Wang, and Xiangyang Ji.
\newblock Cdpn: Coordinates-based disentangled pose network for real-time
  rgb-based 6-dof object pose estimation.
\newblock In {\em Proceedings of the IEEE/CVF International Conference on
  Computer Vision}, pages 7678--7687, 2019.

\bibitem{umap}
L. {McInnes}, J. {Healy}, and J. {Melville}.
\newblock {UMAP: Uniform Manifold Approximation and Projection for Dimension
  Reduction}.
\newblock {\em ArXiv e-prints}, Feb. 2018.

\bibitem{continuous}
Natalia Neverova, David Novotny, Vasil Khalidov, Marc Szafraniec, Patrick
  Labatut, and Andrea Vedaldi.
\newblock Continuous surface embeddings.
\newblock In {\em NIPS}, 2020.

\bibitem{info-nce}
Aaron van~den Oord, Yazhe Li, and Oriol Vinyals.
\newblock Representation learning with contrastive predictive coding.
\newblock {\em arXiv preprint arXiv:1807.03748}, 2018.

\bibitem{pix2pose}
Kiru Park, Timothy Patten, and Markus Vincze.
\newblock Pix2pose: Pixel-wise coordinate regression of objects for 6d pose
  estimation.
\newblock In {\em Proceedings of the IEEE/CVF International Conference on
  Computer Vision}, pages 7668--7677, 2019.

\bibitem{pvnet}
Sida Peng, Yuan Liu, Qixing Huang, Xiaowei Zhou, and Hujun Bao.
\newblock Pvnet: Pixel-wise voting network for 6dof pose estimation.
\newblock In {\em Proceedings of the IEEE/CVF Conference on Computer Vision and
  Pattern Recognition}, pages 4561--4570, 2019.

\bibitem{local-surface-embeddings}
Giorgia Pitteri, Aur{\'e}lie Bugeau, Slobodan Ilic, and Vincent Lepetit.
\newblock 3d object detection and pose estimation of unseen objects in color
  images with local surface embeddings.
\newblock In {\em ACCV}, 2020.

\bibitem{bb8}
Mahdi Rad and Vincent Lepetit.
\newblock Bb8: A scalable, accurate, robust to partial occlusion method for
  predicting the 3d poses of challenging objects without using depth.
\newblock In {\em Proceedings of the IEEE International Conference on Computer
  Vision}, pages 3828--3836, 2017.

\bibitem{low-freq-bias}
Nasim Rahaman, Aristide Baratin, Devansh Arpit, Felix Draxler, Min Lin, Fred
  Hamprecht, Yoshua Bengio, and Aaron Courville.
\newblock On the spectral bias of neural networks.
\newblock In {\em International Conference on Machine Learning}, pages
  5301--5310. PMLR, 2019.

\bibitem{unet}
Olaf Ronneberger, Philipp Fischer, and Thomas Brox.
\newblock U-net: Convolutional networks for biomedical image segmentation.
\newblock In {\em International Conference on Medical image computing and
  computer-assisted intervention}, pages 234--241. Springer, 2015.

\bibitem{dpodv2}
Ivan Shugurov, Sergey Zakharov, and Slobodan Ilic.
\newblock Dpodv2: Dense correspondence-based 6 dof pose estimation.
\newblock {\em IEEE Transactions on Pattern Analysis and Machine Intelligence},
  2021.

\bibitem{siren}
Vincent Sitzmann, Julien Martel, Alexander Bergman, David Lindell, and Gordon
  Wetzstein.
\newblock Implicit neural representations with periodic activation functions.
\newblock {\em Advances in Neural Information Processing Systems}, 33, 2020.

\bibitem{multipath}
Martin Sundermeyer, Maximilian Durner, En~Yen Puang, Zoltan-Csaba Marton,
  Narunas Vaskevicius, Kai~O Arras, and Rudolph Triebel.
\newblock Multi-path learning for object pose estimation across domains.
\newblock In {\em CVPR}, 2020.

\bibitem{implicit-orientation-learning}
Martin Sundermeyer, Zoltan-Csaba Marton, Maximilian Durner, Manuel Brucker, and
  Rudolph Triebel.
\newblock Implicit 3d orientation learning for 6d object detection from rgb
  images.
\newblock In {\em ECCV}, 2018.

\bibitem{vidal}
Joel Vidal, Chyi-Yeu Lin, Xavier Llad{\'o}, and Robert Mart{\'\i}.
\newblock A method for 6d pose estimation of free-form rigid objects using
  point pair features on range data.
\newblock {\em Sensors}, 18(8):2678, 2018.

\bibitem{posecnn}
Yu Xiang, Tanner Schmidt, Venkatraman Narayanan, and Dieter Fox.
\newblock Posecnn: A convolutional neural network for 6d object pose estimation
  in cluttered scenes.
\newblock {\em arXiv preprint arXiv:1711.00199}, 2017.

\end{thebibliography}
